\documentclass[letterpaper]{article} % DO NOT CHANGE THIS
\usepackage{aaai2026}  % DO NOT CHANGE THIS
\usepackage{times}  % DO NOT CHANGE THIS
\usepackage{helvet}  % DO NOT CHANGE THIS
\usepackage{courier}  % DO NOT CHANGE THIS
\usepackage[hyphens]{url}  % DO NOT CHANGE THIS
\usepackage{graphicx} % DO NOT CHANGE THIS
\usepackage{natbib}  % DO NOT CHANGE THIS AND DO NOT ADD ANY OPTIONS TO IT
\usepackage{caption} % DO NOT CHANGE THIS AND DO NOT ADD ANY OPTIONS TO IT
\usepackage{algorithm}
\usepackage{algorithmic}
\usepackage{xcolor} % 如果要背景色
\usepackage{newfloat}
\usepackage{listings}
\usepackage{makecell}
\usepackage{amsmath, amssymb, amsfonts}
\DeclareCaptionStyle{ruled}{labelfont=normalfont,labelsep=colon,strut=off} % DO NOT CHANGE THIS
\floatstyle{ruled}
\newfloat{listing}{tb}{lst}{}
\floatname{listing}{Listing}
\usepackage{amsmath}
\nocopyright 
\title{Enhancing Agentic RL with Progressive Reward Shaping and Value-based Sampling Policy Optimization}
\author {
    Jianghao Su\textsuperscript{\rm 1}\thanks{Also affiliated with Xi'an Jiaotong University, work done during internship at Fliggy Alibaba},
    Xia Zeng\textsuperscript{\rm 1}\thanks{Corresponding Author},
    Luhui Liu\textsuperscript{\rm 1},
    Luo Chao\textsuperscript{\rm 1},
    Ye Chen\textsuperscript{\rm 1},
    Zhuoran Zhuang\textsuperscript{\rm 1}
}
\affiliations {
    \textsuperscript{\rm 1}Fliggy Alibaba \\
    Zhejiang, Hangzhou \\
    jh.su@stu.xjtu.edu.cn, 
    \{zengxia.xz, liuluhui.llh, guotai.lc, nantian, chen.yechen\}@alibaba-inc.com
}
\begin{document}
\maketitle

\begin{abstract}
Large Language Models (LLMs) empowered with Tool-Integrated Reasoning (TIR) can iteratively plan, call external tools, and integrate returned information to solve complex, long-horizon reasoning tasks. Agentic Reinforcement Learning (Agentic RL) optimizes such models over full tool-interaction trajectories, but two key challenges hinder effectiveness:
(1) Sparse, non-instructive rewards, such as binary 0–1 verifiable signals, provide limited guidance for intermediate steps and slow convergence;
(2) Gradient degradation in Group Relative Policy Optimization (GRPO), where identical rewards within a rollout group yield zero advantage, which reducing sample efficiency.

To address these challenges, we propose two complementary techniques: Progressive Reward Shaping (PRS) and Value-based Sampling Policy Optimization (VSPO). PRS is a curriculum-inspired reward design that introduces dense, stage-wise feedback — encouraging models to first master parseable and properly formatted tool calls, then optimize for factual correctness and answer quality. We instantiate PRS for short-form QA (with a length-aware BLEU to fairly score concise answers) and long-form QA (with LLM-as-a-Judge scoring to prevent reward hacking). VSPO is an enhanced GRPO variant that replaces zero advantages samples with prompts selected by a task-value metric balancing difficulty and uncertainty, and applies value-smoothing clipping to stabilize gradient updates.

Experiments on multiple short-form and long-form QA benchmarks show that PRS consistently outperforms traditional binary rewards, and VSPO achieves superior stability, faster convergence, and higher final performance compared to SFT, PPO and GRPO baselines. Together, PRS and VSPO yield LLM-based TIR agents that generalize better across domains.
\end{abstract}

\section{Introduction}
Large language models (LLMs) have shown remarkable capabilities in complex reasoning and tool-augmented decision-making, enabling applications such as question answering, planning, and interactive agents~\cite{zhao2023survey,huang2024understanding}. 
When deployed as Tool-Integrated Reasoning (TIR) agents, LLMs are equipped with external tools --- e.g., search APIs or knowledge bases --- and can iteratively plan, call tools, and integrate their outputs to produce final answers. This multi-step, interleaved reasoning-with-action process is particularly effective in long-horizon tasks such as multi-hop question answering or problem solving in dynamic environments.

Training high-performing LLM-based TIR agents goes beyond single-turn reinforcement learning~\cite{zhang2025landscape}. 
Agentic reinforcement learning (Agentic RL) explicitly optimizes the LLM policy over full tool-interaction trajectories, with feedback signals derived from the correctness and quality of the intermediate and final outputs. 
However, in practice, Agentic RL for TIR faces two key challenges:

\begin{itemize}
    \item \textbf{Sparse and non-instructive rewards:} Widely used reinforcement learning with verifiable rewards (RLVR)~\cite{liu2025reinforcement} often employs binary 0-1 signals to indicate pass/fail based on the final answer. These sparse signals provide limited guidance for improving intermediate tool-use steps, offer no notion of partial progress, and can lead to inefficient exploration and unstable optimization.
    \item \textbf{Low sample efficiency in GRPO:} Group Relative Policy Optimization (GRPO)~\cite{guo2025deepseek} is a strong baseline for multi-sample policy improvement, but when all rollouts in a group receive identical rewards, their computed advantages become zero. This results in dampened gradient updates and low sample efficiency.
\end{itemize}

To address these issues, we propose two complementary techniques for training LLM-based TIR agents:

\begin{itemize}
    \item \emph{Progressive Reward Shaping} (PRS): a curriculum-inspired reward design that augments sparse verifiable rewards with dense intermediate signals, enabling the model to first master essential capabilities --- generating parseable and correctly formatted tool calls --- before optimizing more challenging objectives such as factual correctness and answer quality. In short-form QA, PRS incorporates a length-aware BLEU to avoid unfair penalization of correct short answers; in long-form QA, it integrates LLM-as-a-Judge scoring to mitigate reward hacking.
    \item \emph{Value-based Sampling Policy Optimization} (VSPO): an improved variant of GRPO that replaces low-value (zero-advantage) samples with prompts selected according to a task-value metric balancing uncertainty and difficulty. VSPO further applies value-smoothing clipping to maintain stable gradient scales even when high-value prompts are repeatedly sampled.
\end{itemize}

Experiments on long-form and short-form QA benchmarks demonstrate that PRS consistently outperforms traditional binary reward schemes, and VSPO achieves more stable training and superior performance compared to SFT, PPO and GRPO baselines. Together, PRS and VSPO yield LLM-based TIR agents that converge faster, generalize better across domains.

Our contributions are summarized as follows:
\begin{itemize}
    \item We design PRS, a general reward shaping framework that integrates stage-wise progression and dense intermediate feedback, instantiated for both short- and long-form QA.
    \item We develop VSPO, a GRPO variant with value-based sampling and value smoothing clipping, improving sample efficiency and training stability and performance.
    \item We empirically validate PRS and VSPO across multiple QA benchmarks, showing consistent gains over strong baselines.
\end{itemize}

\begin{figure*}[!ht]
    \centering
    \includegraphics[width=\textwidth]{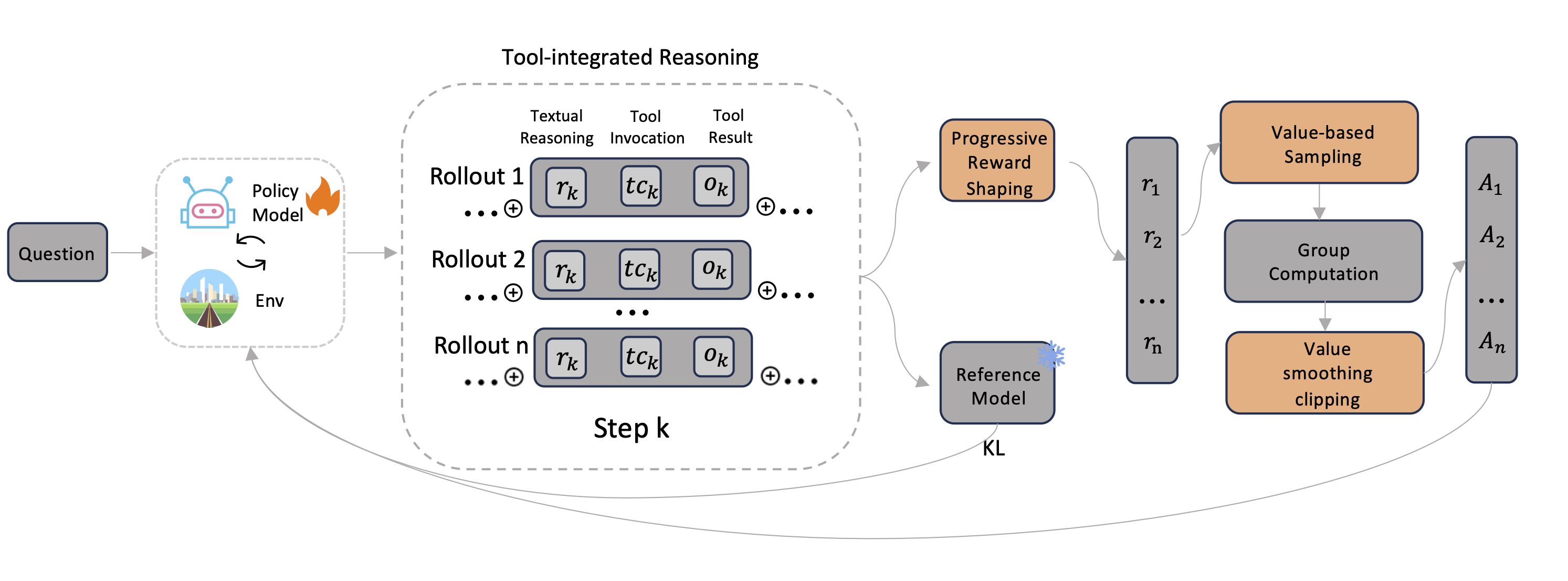}
    \caption{Overview of PRS (rogressive Reward Shaping ) and VSPO(alue-based
Sampling Policy Optimization). }
    \label{fig:overview}
\end{figure*}

\section{Related Work}
% \subsubsection{Retrieval Augmented Generation (RAG). } 
% Retrieval-Augmented Generation (RAG) combines the generative capabilities of large language models (LLMs) with real-time retrieval from external sources to address limitations such as outdated knowledge and hallucinations in static pre-trained models. A typical RAG pipeline consists of three components: retrieval, augmentation, and generation~\cite{leeagent}. Early Naïve RAG systems relied on keyword-based methods (e.g., TF-IDF, BM25) to fetch documents, which limited contextual relevance and scalability~\cite{gao2023retrieval}. Subsequent advances, such as Advanced RAG, incorporated dense retrieval models like DPR~\cite{karpukhin2020dense} and multi-hop reasoning to improve semantic precision. More recent paradigms, including Modular RAG and Graph RAG~\cite{peng2024graph}, introduce hybrid retrieval strategies, reusable pipeline components, and structured knowledge integration to better support complex queries. Despite these advancements, traditional RAG workflows remain largely static and struggle with dynamic multi-step reasoning in real-world settings.

\subsubsection{Tool-Integrated Reasoning.}
Tool-Integrated Reasoning (TIR)~\cite{lin2025understanding} augments LLMs by enabling them to interleave natural-language reasoning with external tool calls, such as code execution or search, thereby improving adaptability in complex tasks. Early TIR systems based on in-context learning or SFT~\cite{yao2022react, schick2023toolformer} achieve promising results but often generalize poorly to unseen tasks and new tool configurations. Recent advances employ Reinforcement Learning with Verifiable Rewards (RLVR)~\cite{liu2025reinforcement}, allowing agents to learn robust tool-use strategies through exploration, outperforming purely imitative methods. RL frameworks such as ToolRL~\cite{qian2025toolrl}, AutoTIR~\cite{wei2025autotir}, and VTool-R1~\cite{wu2025vtool} interleave symbolic computation with language reasoning in unified rollouts, balancing precise tool-mediated operations with flexible verbal inference.

\subsubsection{Agentic RAG. }
Agentic RAG~\cite{ravuru2024agentic} enhances traditional RAG frameworks by embedding autonomous agents that govern the retrieval-generation process, thereby enabling dynamic decision-making, iterative refinement, and adaptive orchestration. Agents leverage established patterns such as \emph{reflection}~\cite{madaan2023self}, \emph{planning}~\cite{madaan2023self}, \emph{tool use}, and \emph{multi-agent collaboration}~\cite{guo2024large} to manage retrieval strategies and integrate diverse information sources. Architectures range from single-agent systems to multi-agent and hierarchical designs, with some frameworks incorporating graph-based reasoning (\emph{Agent-G}~\cite{leeagent}) for richer contextual integration. Compared to static RAG, Agentic RAG achieves greater flexibility, scalability, and contextual accuracy in domains such as healthcare, finance, and education~\cite{singh2024revolutionizing}.

\subsubsection{Training paradigm. }
Post-pretraining optimization of LLMs encompasses Supervised Fine-Tuning (SFT) and Reinforcement Learning (RL). SFT leverages high-quality expert trajectories 
to align model behavior through imitation, but often exhibits limited generalization capability on out-of-distribution inputs or tasks demanding 
novel reasoning patterns not present in the training data. In contrast, RL facilitates active exploration by LLMs, enabling them to identify and reinforce superior behaviors that enhance generalization performance.
RL methods for LLMs can be categorized as critic-based or critic-free. Critic-based approaches like PPO~\cite{schulman2017proximal} and VAPO~\cite{yue2025vapo} use a value function to provide fine-grained feedback 
during training. Critic-free methods like GRPO~\cite{shao2024deepseekmath} and CISPO~\cite{chen2025minimax} use only sequence-level rewards, offering greater computational efficiency and scalability.

%\textbf{Reward Engineering. }

\section{Preliminary}

\subsection{Problem Formalization}
\label{sec:Problem_Formalization}
In this section, we provide a formal defination of TIR (Tool-Integrated Reasoning).

Given a task $x = (h, q)$ where $q$ is the questioin and $h$ is the related background information (e.g., historical dialogue records in dialogue system), the environment $\varepsilon$  provides access to a set of tools $\mathcal{T}  = \{ t_0, t_1, \dots t_n \}$, the language model $\mathcal{M}$ can proactively and iteratively interact with the environment by reasoning and calling specific tools in $\mathcal{T}$, obtaining the corresponding tool results from $\varepsilon$. This process repeats until $\mathcal{M}$ provides the final answer to $x$. This interaction process can be expressed as a multi-step trajectory $\mathbb{T}$.

The steps in $\mathbb{T}$ can be divided into two categories: intermediate tool calls and final answer. The intermediate tool call step $\tau_k$ can be defined as

\begin{equation}
\tau_k = (r_k, tc_k, o_k),
\label{eq:tool_call_step}
\end{equation}

where $r_k$, $tc_k$ and $o_k$ denote the reasoning, tool call and returned observation respectively. The complete trajectory $\mathbb{T}$ is then defined as 

\begin{equation}
\mathbb{T} = \tau_1 \cdot \tau_2 \cdot \dots \tau_n \cdot \tau_{\text{final}},
\label{eq:multi_step_trajectory}
\end{equation}

where$\tau_{\text{final}} = (r_{\text{final}}, a)$ and $a$ denotes the final answer generated by $\mathcal{M}$.

\subsection{Agentic Reinforcement Learning}
Based on Section~\ref{sec:Problem_Formalization}, the Agentic RL training objective is:
\begin{multline}
\max_{\pi_{\theta}} 
\mathbb{E}_{q \sim \mathcal{D}, \mathbb{T} \sim \pi_{\theta}(\cdot \mid x ; \mathcal{E})}
\left[r_{\phi}(q, \mathbb{T})\right]
\\
- \beta \mathbb{D}_{kl}\left[
\pi_{\theta}(\mathbb{T} \mid x ; \mathcal{E}) \| 
\pi_{\mathrm{ref}}(\mathbb{T} \mid x ; \mathcal{E})
\right],
\label{eq:general_rl}
\end{multline}

where $\mathbb{T}$ denotes the trajectory generated by $\mathcal{M}$, $\pi_{\theta}$, and $\pi_{ref}$ denotes the policy model and reference model, respectively. $r_{\phi}$ and $D_{kl}$ denotes the returns and KL divergence respectively.

There are two well-established policy-gradient RL methods used in Agentic RL, Proximal Policy Optimization (PPO) \cite{schulman2017proximal} and Group Relative Policy Optimization (GRPO) \cite{guo2025deepseek}.

For PPO, based on the form of  \ref{eq:general_rl}, the $r_{\phi}(q, \mathbb{T})$ of PPO is 
\begin{align}
r_{\phi}(q, \mathbb{T})
&= \frac{1}{N} \sum_{i=1}^{N} 
\min \left(
\frac{\pi_{\theta}\left(a_{t}^{(i)} \mid s_{t}\right)}
     {\pi_{\theta_{\text{old}}}\left(a_{t}^{(i)} \mid s_{t}\right)}
A\left(s_{t}, a_{t}^{(i)}\right), \right. \nonumber \\
&\quad\left.
\operatorname{clip}\left(
\frac{\pi_{\theta}\left(a_{t}^{(i)} \mid s_{t}\right)}
     {\pi_{\theta_{\text{old}}}\left(a_{t}^{(i)} \mid s_{t}\right)},
1-\epsilon, 1+\epsilon
\right) 
A\left(s_{t}, a_{t}^{(i)}\right)
\right),
\end{align}
where $a_{t}^{(i)} \sim \pi_{\theta_{\text {old }}}\left(a \mid s_{t}\right)$ is the i-th sampled token from the old policy $\pi_{\theta_{\text {old }}}$. $A_t$ is the estimated advantage given by
\begin{equation}
A\left(s_{t}, a_{t}\right)=\mathcal{R}\left(s_{t}, a_{t}\right)-V\left(s_{t}\right),
\end{equation}
where $V_{\theta}(s)$ is the learned value function, which is derived from a  critic network.

For GRPO, the $r_{\phi}(q, \mathbb{T})$ of GRPO is 
\begin{equation}
\begin{aligned}
&\mathcal{J}_{\mathrm{GRPO}}(\theta)
= \mathbb{E}_{x \sim \mathcal{D},\ \{y_{i}\}_{i=1}^{G} \sim \pi_{\theta_{\text{old}}}(\cdot \mid x)} \\
&\quad \left[ \frac{1}{G} 
    \sum_{i=1}^{G} \frac{1}{|y_{i}|} 
    \sum_{t=1}^{|y_{i}|} 
\right. \\
&\quad\quad \left.
    \min\big( w_{i,t}(\theta) \hat{A}_{i,t},\ 
    \operatorname{clip}\big(w_{i,t}(\theta), 1-\epsilon, 1+\epsilon\big)\hat{A}_{i,t} \big)
\right], \\
&w_{i,t}(\theta)
= \frac{\pi_{\theta}\left(y_{i,t} \mid x, y_{i,<t}\right)}
        {\pi_{\theta_{\text{old}}}\left(y_{i,t} \mid x, y_{i,<t}\right)}, \\
&\hat{A}_{i,t} 
= \hat{A}_{i} 
= \frac{R\left(x, y_{i}\right) - \operatorname{mean}\big(\{R(x, y_{j})\}_{j=1}^{G}\big)}
        {\operatorname{std}\big(\{R(x, y_{j})\}_{j=1}^{G}\big)} .
\end{aligned}
\label{eq:grpo_loss}
\end{equation}

Here $x_i$ is the task (A.K.A prompt) and $y_i$ is the multi-step trajectory (A.K.A rollout) defined is equation \ref{eq:multi_step_trajectory}. All tokens in $y_i$ share the same advantage as $\hat{A}_i$.

\subsection{Reward Shaping}
Reward shaping belongs to the category of reward design, which is a reinforcement learning (RL) technique that augments the original reward function with additional signals to accelerate and guide the agent’s learning process, while preserving the optimal policy\cite{ibrahim2024comprehensive}. In many real-world RL tasks, rewards are sparse, delayed, or noisy, making it difficult for agents to effectively explore the environment and converge to desirable behaviors. Reward shaping addresses these challenges by incorporating additional incentives or penalties that provide intermediate feedback, thereby encouraging exploration toward promising states and actions.

\section{Methodology}
Our methodology consists of two components: Progressive Reward Shaping and Value-based Sampling.

\subsection{Progressive Reward Shaping (PRS)}\label{sec:reward shaping}
We proposed progressive reward shaping (PRS) based on the following  key insights: a effective reward function should exhibit (1) richer partial ordering and (2) instructive progression, both of which can significantly improve learning efficiency.

\textbf{Richer partial order.} A reward function with richer partial ordering—also described as being more discriminative or dense—can distinguish between rollouts of similar quality rather than assigning identical rewards.  For example, the commonly used Reinforcement Learning with Verifiable Rewards (RLVR)~\cite{liu2025reinforcement} typically employs a binary reward of 0 and 1. Such sparse binary rewards provide limited learning signals: the model only receives "correct" or "incorrect" feedback, without finer-grained "better" or "worse" signals to guide improvement. Furthermore, for GRPO, this sparsity increases the likelihood of generating  zero-advantage samples.

\textbf{More instructive.} Inspired by curriculum Learning~\cite{soviany2022curriculum}, we design the reward function to start with simple, fundamental rewards and progressively transition into more challenging criteria. This staged approach encourages the model to first learn basic behaviors before advancing to more complex reasoning patterns, facilitating more stable and efficient learning.

Based on these principles, we present reward design framework in retrieval-augmented question-answering scenarios. Without loss of generality, this reward framework is applicable to  diverse agentic RL training scenarios, and we provide a generalized formulation that can be adapted to other domains.

\subsubsection{Short-Form QA}
Short-Form QA refers to a type of QA where the answers are typically brief and mostly consist of concrete entities (e.g. time, place and name).

The standard reward function of short-form QA is binary 0-1 reward based on EM (Exact Match)~\cite{jin2025search}:
\begin{equation}
r_{\phi}(x, y)=\operatorname{EM}\left(a_{\text {pred }}, a_{\text {gold }}\right)
\label{eq:em}
\end{equation}
where $a_{\text {pred}}$ is the extracted final answer from response $y$ and $a_{\text{gold}}$ is the ground truth answer.

For TIR, model $\mathcal{M}$ must generate the correct intermediate tool call step $\tau_k$ of $\mathbb{T}$ before producing the final answer step $\tau_{\text{final}}$. We identify three progressive learning objectives:
\begin{enumerate}
    \item Complete process: each tool call $tc_k$ in $\tau_k$ can be successfully parsed and executed to return observations $o_k$ and the final step $\tau_{\text{final}}$ can be parsed to extract the final answer $a$.
    \item Complete format: the content generated at each step to be correctly placed in the corresponding tags (e.g. \texttt{<reasoning></reasoning>})
    \item Correct answer: based on $\bigcup_{i}o_i$, $\mathcal{M}$ generates the correct answer $a$.
\end{enumerate}
We design corresponding reward components for each learning objective. We will first introduce each reward and then provide the complete PRS formulation.

\textbf{Process reward $R_{process}$.} The definition of $R_{process}$ is:
\begin{equation}
R_{\text{process}} =
\begin{cases}
    -1 & \text{other}, \\
    0  & \forall tc_k \in \mathbb{T} \ \text{can be parsed but } a \ \text{cannot be parsed}, \\
    1  & \forall tc_k, a \in \mathbb{T},\ tc_k \ \text{and} \ a \ \text{can be correctly parsed}
\end{cases}
\label{eq:process}
\end{equation}

Process reward evaluates if $\mathcal{M}$ can generate parseable tool calls and final answers.

\textbf{Format reward $R_{format}$.} The defination of $R_{format}$ is:
\begin{equation}
R_{format} = \\
\begin{cases}
    0 & \text{others} \\
    0.1 & \text{Complete label package} 
\end{cases}
\label{eq:format}
\end{equation}
The format reward evaluates whether $\mathcal{M}$ generates outputs that conforms to the required structured format.

\textbf{Answer reward $R_{a}$.} 
Unlike the binary EM reward in Eq.~\eqref{eq:em}, we design a denser reward based on BLEU ~\cite{papineni2002bleu}, named short-form BLEU, which is more suitable for short-form QA.

BLEU is a widely-used metric for machine translation evaluation. It measures the n-gram overlap between a predicted translation and one or more reference texts. The definition of BLEU is 
\begin{equation}
\begin{aligned}
\mathrm{BLEU} &= \mathrm{BP} \cdot \exp\left(\sum_{n=1}^{N} w_{n} \log p_{n}\right), \\
\mathrm{BP} &=
\begin{cases}
1, & \text{if } c > r, \\
\exp\left(1 - \frac{r}{c}\right), & \text{if } c \le r.
\end{cases}
\end{aligned}
\label{eq:bleu}
\end{equation}

where $N=4$, $p_n$ is the modified $n$-gram precision, $w_n$ is the weight for each $n$-gram (usually $1/N$), 
$c$ is the candidate length, $r$ is the reference length, and $\mathrm{BP}$ is the brevity penalty.

However, standard BLEU with $N=4$ is problematic for short-form QA. When the predicted answer length is less than 4 tokens, even an exact match results in BLEU $< 1$ due to undefined higher-order $n$-gram precisions. This artificially penalizes short answers and biases the model toward generating longer responses.

To address this issue, we propose \textit{short-form BLEU}, which dynamically adjusts the $n$-gram weights $w_n$ according to the prediction length $c$. Specifically, let
\begin{equation}
\max_n = \min(N, c),
\end{equation}
then we re-define the weights as
\begin{equation}
w_n =
\begin{cases}
\frac{1}{\max_n}, & n \le \max_n ,\\
0, & n > \max_n
\end{cases}
\end{equation}

so that only valid $n$-grams up to $\max\_n$ contribute equally to the final score.
This ensures that an exact match between a short predicted answer and the reference can achieve $\mathrm{BLEU} = 1$, avoiding unintended reward degradation.

\textbf{Complete Formula of PRS.}  
The defination of short-formal PRS is
\begin{equation}
PRS_{short} = \\
\begin{cases}
    R_{process} + R_{format} & \text{ if } R_{process} < 1\\
    R_{process} + R_{format} + R_{a} & \text{others} 
\end{cases}
\label{eq:prs_short}
\end{equation}

Note that $R_{process} + R_{format} + R_{a} \ge R_{process} + R_{format}$, so PRS encourage models to first have the ability to conduct complete interactions, and then improve the quality of final answer.

\subsubsection{Long-Form QA}
Different from short-form QA, relying solely on n-gram based metrics as answer reward is insufficient for long-form responses, as models exploit these metrics through reward hacking. To overcome this issue, we add a reward of LLM-as-a-Judge $R_{judge}$ on top of $PRS_{short}$. $R_{judge}$ mainly evaluates whether the model's answers are factually grounded and free from hallucinations. The defination of $PRS_{long}$ is
\begin{equation}
\begin{split}
& PRS_{long} = \\
& \begin{cases}
    R_{process} + R_{format}, & \text{if } R_{process} < 1,\\
    R_{process} + R_{format} + R_{judge}, & \text{if } R_{process} \ge 1,\\
    R_{process} + R_{format} + R_{judge} + R_{a}, & 
    \begin{aligned}
         &\text{if } R_{process} \ge 1 \\
         &\text{and } R_{judge} \ge 1
    \end{aligned}
\end{cases}
\end{split}
\end{equation}

$PRS_{long}$ encourage models to first generate a reasonable answer without illusions, and then maximize alignment with the reference answer.

The selection of reward weights is based on experience. $R_{process}$, $R_{judge}$ and $R_{a}$ are equal weights and $R_{format}$ is lower weight cause format not as important as the first three.

\subsubsection{The general form of PRS}
PRS is not limited to QA scenarios and can serve as a design framework for various agentic RL training tasks. While $R_{\text{process}}$ and $R_{\text{format}}$ are universal components in TIR training, the framework can be extended with task-specific reward components. Let $\sigma(\cdot)$ be the sigmoid function, then the general form of PRS would be 
\begin{equation}
\begin{split}
R_{\mathrm{PRS}} = & R_1 + \mathbb{I}\!\left( R_1 \ge \epsilon_1 \right) \cdot \sigma(R_2)\\
&  + \mathbb{I}\!\left( R_1 \ge \epsilon_1, R_2 \ge \epsilon_2 \right) \sigma(\cdot R_3) + \cdots.
\end{split}
\end{equation}
Where $\epsilon_i$ is the reward threshold to control reaching the next reward step. The sigmoid function $\sigma(\cdot)$ ensures that the cumulative reward at each stage is strictly greater than at the previous stage (i.e., $\sigma(R_i) \in (0,1)$), thereby encouraging the model to progressively learn increasingly complex behaviors while maintaining stable gradient signals.

\subsection{VSPO (Value-based Sampling Policy Gradient)} \label{sec:Value-based Sampling}
GRPO suffers from the gradient degradation problem
 when all rollouts within their group have the same reward. The resulting advantages are zero, leading to zero policy gradients. This has a negative impacts:
\begin{itemize}
    \item the gradient signal of the batch is dampened, resulting in a decrease in sample efficiency.
\end{itemize}
To address this issue, we propose Value-based Sampling Policy Optimization (VSPO),  an improved variant of GRPO. Unlike the dynamic sampling approach in DAPO~\cite{yu2025dapo}, which may be computationally intensive. VSPO replaces zero advantage samples with selectively sampling in the current batch, maintaining computational efficiency while improving sample utilization and model learning efficiency.

\subsubsection{The introduction of VSPO}
To ensure all samples in the batch produce effective gradients, 
VSPO first filters out the tasks $x$ in the current batch with reward variance 
$Var\!\left(\{ R(x, y_{i}) \}_{i=1}^{G}\right)$ less than the threshold $\varepsilon$ 
(here we set $\varepsilon$ as $1\mathrm{e}{-}6$), 
then replaces them by resampling from the remaining samples.

Denote current batch as $B$ and the rollout numbers is $n$, $\forall x = (h, q) \in B$, . the rollout group of $x$ is $G_{x} = \{(y_i, r_i)\}_{i = 1}^{n}$, here $y_i$ denotes the rollout and $r_i$ is the corresponding reward. Let $\mu_{x}$ and $\sigma_{x}^2$ be the mean and variance of rewards within $G_{x}$. That is 
\begin{equation}
\mu_{x} = \frac{1}{n} \sum_{i=1}^{n} r_i
\label{eq:mu_x}
\end{equation}

\begin{equation}
\sigma_{x}^2 = \frac{1}{n} \sum_{i=1}^{n} \left(r_i - \mu_x\right)^2
\label{eq:sigma_x}
\end{equation}
After filters out the tasks $x$ in the current batch with reward variance $\sigma_{x}^2$ less than the threshold $\varepsilon$, we have $B = B^{\sigma_{x}^2 >= \varepsilon } \cup B^{\sigma_{x}^2 < \varepsilon }$ and $B^{\sigma_{x}^2 >= \varepsilon } \cap B^{\sigma_{x}^2 < \varepsilon } = \emptyset $.

A naive approach would be uniform sampling from $B^{\sigma_{x}^2 >= \varepsilon }$. However, we argue this can be improved. The insight of VSPO is "prioritize samples with higher learning value tasks".

% During the training process of GRPO, there are two types of tasks are prone to zero advantages:

% \begin{itemize}
%     \item Too easy task: all rollouts of the task  achieve the same high reward, indicating the model $\mathcal{M}$ already proficient in solving this task.
%     \item Too hard task: all rollouts receive the same low reward, which means model $\mathcal{M}$ lacks the necessary capability.
% \end{itemize}
\textbf{The learning value of task.}
To measure the value of task $x$, we reinterpret the reward mean $\mu_x$ and varience $\sigma^2_x$ of rollouts group $G_x$ in GRPO. 
\begin{itemize}
\item $\mu_x$ implies the average ability for $\mathcal{M}$ to solve $x$. The higher the $\mu_x$, the better $x$ is solved by $\mathcal{M}$. 
\item $\sigma_{x}^2$ reflects the policy's uncertainty in solving $x$ or the diversity of exploration. The higher the $\sigma^2_x$, the more rollouts $G_x$ contains with relatively high and low rewards, that is to say, more gradient signals it contains.
\end{itemize}

We believe that the more difficult it is for $\mathcal{M}$ to solve and the more gradient signals the task contains have richer learning value. Based on the above insights, we define the value score of task $x$ as:
\begin{equation}
V_x = (R_{max} - \mu_{x}) \cdot \sigma^2_x,
\label{eq: value}
\end{equation}
where $R_{max} = \max_{x \in B, y_i \in G_{x}} R(x, y_i)$ is the maximum reward of current batch $B$. $\mu_{x}$ and $\sigma^2_x$ is defined in equation \ref{eq:mu_x} and \ref{eq:sigma_x}.

The term $(R_{\max} - \mu_x)$ captures the difficulty of $x$ and represents the potential improvement gap between the best rollout and average performance, while $\sigma_x^2$ captures the policy's uncertainty. This formulation balances both difficulty and uncertainty, ensuring that the more difficult for $\mathcal{M}$ to solve and the more gradient signals the task contains have higher learning value.

\textbf{Value-based sampling} 
Based on eaquation \ref{eq: value}, we normalize the values using a temperature-controlled Softmax:
\begin{equation}
    p_{x_i} = \frac{\exp(V_{x_i} / T)}{\sum_{k} \exp(V_{x_i} / T)}, \quad T>0,
    \label{eq:tmp_sample}
\end{equation}
where smaller $T$ increases preference for high value score samples, 
while a larger $T$ yields more uniform probabilities.

Finally, we sample $\{x_i\}$ from $B^{\sigma^2 >= \varepsilon}$ according to $\{p_{x_i}\}$ to replace the filtered samples. 

Handling of some boundary situations, set current batch is $B$: 
\begin{itemize}
    \item If $\forall x \in B$ have $\sigma_{x}^2 > \epsilon$, which means there is no zero advantage task, then value-based sampling will not be performed. The processing flow is exactly the same as traditional GRPO.
    \item IF $\forall x \in B$ have $\sigma_{x}^2 <= \epsilon$, which means every task of $B$ is zero advantage, then value-based sampling will not be performed, too. The processing flow is exactly the same as traditional GRPO.
\end{itemize}

\textbf{Value smoothing clipping}
Value-based sampling alone can cause training instability. If prompt $x_K$ is sampled $N$ times, its contribution to the objective in Eq.~\eqref{eq:grpo_loss} becomes:

\begin{align}
\mathcal{J}_{x_K}(\theta) 
&= N \!\times\!
\sum_{\{y_{i}\}_{i=1}^{G} \!\sim\! \pi_{\theta_{\text{old}}}(\cdot \mid x_K)}
\Bigg[
    \frac{1}{G} \sum_{i=1}^{G} 
    \frac{1}{|y_{i}|} 
\nonumber \\
&\quad\times \sum_{t=1}^{|y_{i}|}
    \min\!\Big(
        w_{i,t}(\theta) \, \hat{A}_{i,t}, 
\nonumber \\
&\quad\quad
        \operatorname{clip}\!\big(w_{i,t}(\theta),\, 1-\epsilon,\, 1+\epsilon\big) 
        \hat{A}_{i,t}
    \Big)
\Bigg]
\nonumber \\
&= \sum_{\{y_{i}\}_{i=1}^{G} \!\sim\! \pi_{\theta_{\text{old}}}(\cdot \mid x_K)}
\Bigg[
    \frac{1}{G} \sum_{i=1}^{G} 
    \frac{1}{|y_{i}|} 
\nonumber \\
&\quad\times \sum_{t=1}^{|y_{i}|}
    \min\!\Big(
        w_{i,t}(\theta) \cdot N \cdot \hat{A}_{i,t}, 
\nonumber \\
&\quad\quad
        \operatorname{clip}\!\big(w_{i,t}(\theta),\, 1-\epsilon,\, 1+\epsilon\big)
        \cdot N \cdot \hat{A}_{i,t}
    \Big)
\Bigg]
\label{eq:grpo_loss}
\end{align}

This shows that sampling $x_K$ $N$ times is equivalent to scaling its advantage $\hat{A}_i$ by $N$, which can destabilize policy gradients.

To stabilize training, we further proposed value smoothing clipping:

\begin{equation}
A_{\text{new}} = \left(\alpha - \frac{\alpha - 1}{N}\right) \cdot A,
\end{equation}
where $\alpha \in [1, +\infty]$ controls the the clipping magnitude (note that $\lim_{N \to + \infty} (\alpha - \frac{\alpha - 1}{N}) = \alpha$)
When $N=1$, $A_{\text{new}} = A$; when $N>1$, $A$ is smoothly reduced to 
prevent over-dominance of repeated samples in gradient updates.

\subsubsection{A Dynamic Advantage Weight Perspective of VSPO}
VSPO can be interpreted as task-value-based reweighting rather than random sampling. Consider the general policy gradient~\cite{sutton1999policy} with advantage function $A_t$:
\begin{equation}
\nabla_{\theta} \mathcal{J}(\theta)=\mathbb{E}_{x \sim \mathcal{D}, y \sim \pi_{\theta}}\left[\sum_{t=1}^{T} \nabla_{\theta} \pi_{\theta}\left(y_{t} \mid y_{<t}\right) A_{t}\right] .
\end{equation}
In GRPO, all tokens in the same trajectory share the same advantage, so:
\begin{equation}
\nabla_{\theta} \mathcal{J}(\theta)=\mathbb{E}_{x \sim \mathcal{D}, y \sim \pi_{\theta}}\left[\sum_{t=1}^{T} \nabla_{\theta} \pi_{\theta}\left(y_{t} \mid y_{<t}\right) \right] \cdot A_{x}.
\end{equation}
Let $\omega = \left(\alpha - \frac{\alpha - 1}{N}\right)$, then
\begin{equation}
\nabla_{\theta} \mathcal{J}(\theta)=\mathbb{E}_{x \sim \mathcal{D}, y \sim \pi_{\theta}}\left[\sum_{t=1}^{T} \nabla_{\theta} \pi_{\theta}\left(y_{t} \mid y_{<t}\right) \right] \cdot \omega_x \cdot A_{x}.
\end{equation}
Thus, tasks with higher learning value receive larger advantage weights $\omega_x \cdot A_x$, effectively increasing the update step size for these tasks and amplifying policies with greater relative advantages.

\section{Experimental Setup} \label{experiments}
To evaluate the effectiveness of PRS and VSPO in training LLM-based TIR agent, we conduct experiments on the following two types of long-horizon reasoning tasks: short-form QA and long-form QA. In these tasks, the agent has access to at least one information retrieval tool to answer the question.

In short-form QA, we use Qwen2.5-3B-Instruct~\cite{qwen2} and in long-form QA, we use the Qwen3-14B~\cite{qwen3technicalreport} as the training model. All of them uses full parameter training. 

We adopt Trinity-RFT~\cite{pan2025trinity} as our training framework.

For VSPO, we set $T=0.1$ and $\alpha = 2$. The rollout num of VSPO, CISPO and GRPO is 5. The learning rate of all algorithms is 1e-7. Training is performed on a single node with 8 H20 GPUs (141GB). To optimize GPU memory usage, we enable gradient checkpointing and use Fully Sharded Data Parallel (FSDP) with CPU offloading.

We also use gradient checkpointing, FSDP offloading, and vLLM-based rollouts with the same hyperparameters as above. The rollout temperature and top-p values are both set to 1.0, and the KL divergence coefficient $\beta$ and clip ratio $\epsilon$ are fixed at 0.001 and 0.2.

\subsection{Datasets} 
\subsubsection{Long-Form QA. }We use proprietary data from our production system.  We select conversation records from the online conversation database as RL rollout data. There are three types of queries in total:
\begin{itemize}
    \item $Q_{simple}$: A plain text query containing a single question.
    \item $Q_{multim}$: A multimodal query containing both images and text.
    \item $Q_{multiq}$: A query containing multiple sub-questions.
\end{itemize}
The number of training and testing sets is 1678 and 334, respectively. We manually annotate reference answers based on our internal knowledge base.

For expert data used in SFT, we use knowledge distillation from the Qwen3-235B-A22B~\cite{qwen3technicalreport} to construct high-quality trajectory data. Specifically, we collect online service logs from Qwen3-235B-A22B and apply rule-based filtering and LLM-as-a-Judge cleaning to produce expert demonstrations.

\subsubsection{Short-Form QA. } Following \cite{jin2025search}, we use seven benchmark datasets: (1) General Question Answering: NQ~\cite{kwiatkowski2019natural}, TriviaQA~\cite{joshi2017triviaqa}, and PopQA~\cite{mallen2023not}. (2) Multi-Hop Question Answering: HotpotQA~\cite{yang2018hotpotqa}, 2WikiMultiHopQA~\cite{ho2020constructing}, Musique~\cite{trivedi2022musique}, and Bamboogle~\cite{press2023measuring} as Short-Form QA train and test data.

We merge the training sets of NQ and HotpotQA as our training set and evaluation is conducted on the test or validation sets of all seven datasets to assess both in-domain and out-of-domain performance.

\subsection{Evaluation Metrics}
\subsubsection{Long-Form QA.}
We use GPT-4.1 as an LLM judge to determine whether the responses contains hallucinations and whether it matches the annotated reference answers. The evaluation prompts are provided in Appendix \ref{sec:judge_pt}.

\subsubsection{Short-Form QA.} 
We use \texttt{Exact Match} (EM) as the evaluation metric, computed against the golden answers provided by each benchmark dataset.

\subsection{Baselines}
To verify the effectiveness of PRS, we compare it with rule-based 0-1 reward to demonstrate the effectiveness of PRS. All reward functions use GRPO as the RL algorithm. For long-form QA, the 0-1 reward use Qwen3-235B-A22B-Instruct-2507~\cite{qwen3technicalreport} as the verifier, determine whether the model output passes based on the annotated reference answers, if passed, the reward is 1, otherwise it is 0.
For short-form QA, using exact match as rule-based 0-1 reward.

To verify the effectiveness of VSPO, we compare it with GRPO, PPO~\cite{schulman2017equivalence} and SFT-only. All RL algorithms use the same PRS reward.

\section{Main Results}
\label{sec:results}
\subsection{The main results of PRS}
The main results comparing PRS with 0-1 reward are presented in Table \ref{tab:main_prs_long} for long-form QA and Table \ref{tab:main_prs_short} for short-form QA. Under the same RL algorithm, we can found that compared to 0-1 reward, PRS achieved better results. 
In long-form OA, $PRS_{long}$ has increased reference answer pass rate by $0.62\%$ and hallucination Pass Rate by $1.7\%$ compared to 0-1 reward.
In short-form OA, $PRS_{short}$ has increased exact match score by $2.74\%$ compared to 0-1 reward.

\begin{table}[tbp] % 这里改为 table 而不是 table*
\centering
\setlength{\tabcolsep}{4pt} % 调整列间距（可选）
\renewcommand{\arraystretch}{1.2} % 调整行高（可选）
\begin{tabular}{l c c c c}
\hline
\multicolumn{5}{c}{\textbf{Qwen3-14B}} \\
Methods & $Q_{simple}$ & $Q_{multiq}$ & $Q_{multim}$ &Avg. \\ 
\hline
\multicolumn{5}{c}{\textbf{Reference Answer Pass Rate}} \\
Untrained & 0.5513 & 0.6999 & 0.5000 & 0.5897 \\ 
GRPO + 0-1 reward & 0.6955 & 0.8017 & 0.6000 & 0.7255 \\ 
GRPO + $PRS_{long}$        & 0.7144 & 0.8183 & 0.7150 & 0.7300 \\
\hline
\multicolumn{5}{c}{\textbf{Hallucination Pass Rate}} \\
Untrained         & 0.7821 & 0.8167 & 0.7143 & 0.7893 \\ 
GRPO + 0-1 reward & 0.8782 & 0.9333 & 0.8600 & 0.8925 \\ 
GRPO + $PRS_{long}$        & 0.8982 & 0.9333 & 0.9000 & 0.9077\\
\hline
\end{tabular}
\caption{Results on the test set of 0-1 reward and PRS for long-form QA.}
\label{tab:main_prs_long}
\end{table}

\begin{table*}[tbp] % 跨双栏，如果不需要可改为 table
\centering
\setlength{\tabcolsep}{4pt} % 调整列间距（可选）
\renewcommand{\arraystretch}{1.2} % 调整行高（可选）
\begin{tabular}{c c c c c c c c c} % 9列，居中对齐
\hline
\multicolumn{9}{c}{\textbf{Qwen2.5-3B-Instruct}} \\
Methods & NQ & TriviaQA & PopQA & HotpotQA & 2wiki & Musique & Bamboogle & Avg. \\ \hline
\multicolumn{9}{c}{\textbf{Exact Match (EM)}} \\
Untrained & 0.0415 & 0.1751 & 0.0849 & 0.0622 & 0.0891 & 0.0082 & 0.1667 & 0.0960 \\ 
GRPO + 0-1 reward & 0.3213 & 0.5022 & 0.3457 & 0.3000 & 0.2586 & 0.1120 & 0.0833 & 0.3390 \\ 
GRPO + $PRS_{short}$ & 0.3518 & 0.4987 & 0.3562 & 0.3100 & 0.2600 & 0.1245 & 0.3333 & 0.3483 \\ \hline
\end{tabular}
\caption{Results on the test set of 0-1 reward and PRS for short-form QA.}
\label{tab:main_prs_short}
\end{table*}

\subsection{The main results of VSPO}
The main results comparing VSPO with baseline methods across are presented in Table \ref{tab:main_vspo_long} for long-form QA and Table \ref{tab:main_vspo_short} for short-form QA. Under the same reward function, we can found that compared to other RL algorithm, VSPO achieved better results. 
In long-form QA, VSPO has increased reference answer pass rate by $5.34\%$ compared to GRPO.
In short-form OA, $PRS_{short}$ has increased exact match score by $3.16\%$ compared to GRPO.

\begin{table}[tbp] % 这里改为 table 而不是 table*
\centering
\setlength{\tabcolsep}{4pt} % 调整列间距（可选）
\renewcommand{\arraystretch}{1.2} % 调整行高（可选）
\begin{tabular}{l c c c c}
\hline
\multicolumn{5}{c}{\textbf{Qwen3-14B}} \\
Methods & $Q_{simple}$ & $Q_{multiq}$ & $Q_{multim}$ &Avg. \\ 
\hline
\multicolumn{5}{c}{\textbf{Reference Answer Pass Rate}} \\
Untrained & 0.5513 & 0.6999 & 0.5000 & 0.5897 \\ 
SFT & 0.6405 & 0.7500 & 0.6000 & 0.6687 \\ 
GRPO + $PRS_{long}$ & 0.7144 & 0.8183 & 0.7150 & 0.7300 \\
PPO + $PRS_{long}$        & 0.700 & 0.8213 & 0.7020 & 0.7327 \\
VSPO + $PRS_{long}$       & 0.7147  & 0.9000  & 0.8557  & 0.7690  \\
\hline
\multicolumn{5}{c}{\textbf{Hallucination Pass Rate}} \\
Untrained         & 0.7821 & 0.8167 & 0.7143 & 0.7893 \\ 
SFT & 0.8079 & 0.8167 & 0.7556 & 0.8086 \\ 
GRPO + $PRS_{long}$ & 0.8982 & 0.9333 & 0.9000 & 0.9077 \\ 
PPO + $PRS_{long}$        & 0.8282 & 0.8633 & 0.9000 & 0.8399 \\
VSPO + $PRS_{long}$       & 0.8662  & 0.9000  & 0.8571  & 0.8750  \\
\hline
\end{tabular}
\caption{Results on the test set of RL algorithm for long-form QA.}
\label{tab:main_vspo_long}
\end{table}

\begin{table*}[tbp] % 跨双栏，如果不需要可改为 table
\centering
\setlength{\tabcolsep}{4pt} % 调整列间距（可选）
\renewcommand{\arraystretch}{1.2} % 调整行高（可选）
\begin{tabular}{c c c c c c c c c} % 9列，居中对齐
\hline
\multicolumn{9}{c}{\textbf{Qwen2.5-3B-Instruct}} \\
Methods & NQ & TriviaQA & PopQA & HotpotQA & 2wiki & Musique & Bamboogle & Avg. \\ \hline
\multicolumn{9}{c}{\textbf{Exact Match (EM)}} \\
Untrained & 0.0415 & 0.1751 & 0.0849 & 0.0622 & 0.0891 & 0.0082 & 0.1667 & 0.0960 \\ 
SFT & 0.2490 & 0.2920 & 0.1040 & 0.1860 & 0.2480 & 0.0440 & 0.1120 & 0.1760 \\ 
GRPO + $PRS_{short}$ & 0.3518 & 0.4987 & 0.3562 & 0.3100 & 0.2600 & 0.1245 & 0.3333 & 0.3483 \\ 
PPO + $PRS_{short}$ & 0.3210 & 0.4887 & 0.3562 & 0.3200 & 0.2650 & 0.1200 & 0.3325 & 0.3333 \\ 
VSPO + $PRS_{short}$ & 0.3600 & 0.5000 & 0.3634 & 0.3100 & 0.2650 & 0.1367 & 0.3489 & 0.3593 \\ 
\hline
\end{tabular}
\caption{Results on the test set of RL algorithm for short-form QA.}
\label{tab:main_vspo_short}
\end{table*}

\section{Analysis}
We analyze some issues based on long-form QA and short-form QA.

\subsection{VSPO v.s. Other RL Algorithms}
\begin{figure}[htb]
    \centering
    \includegraphics[width=0.9\linewidth]{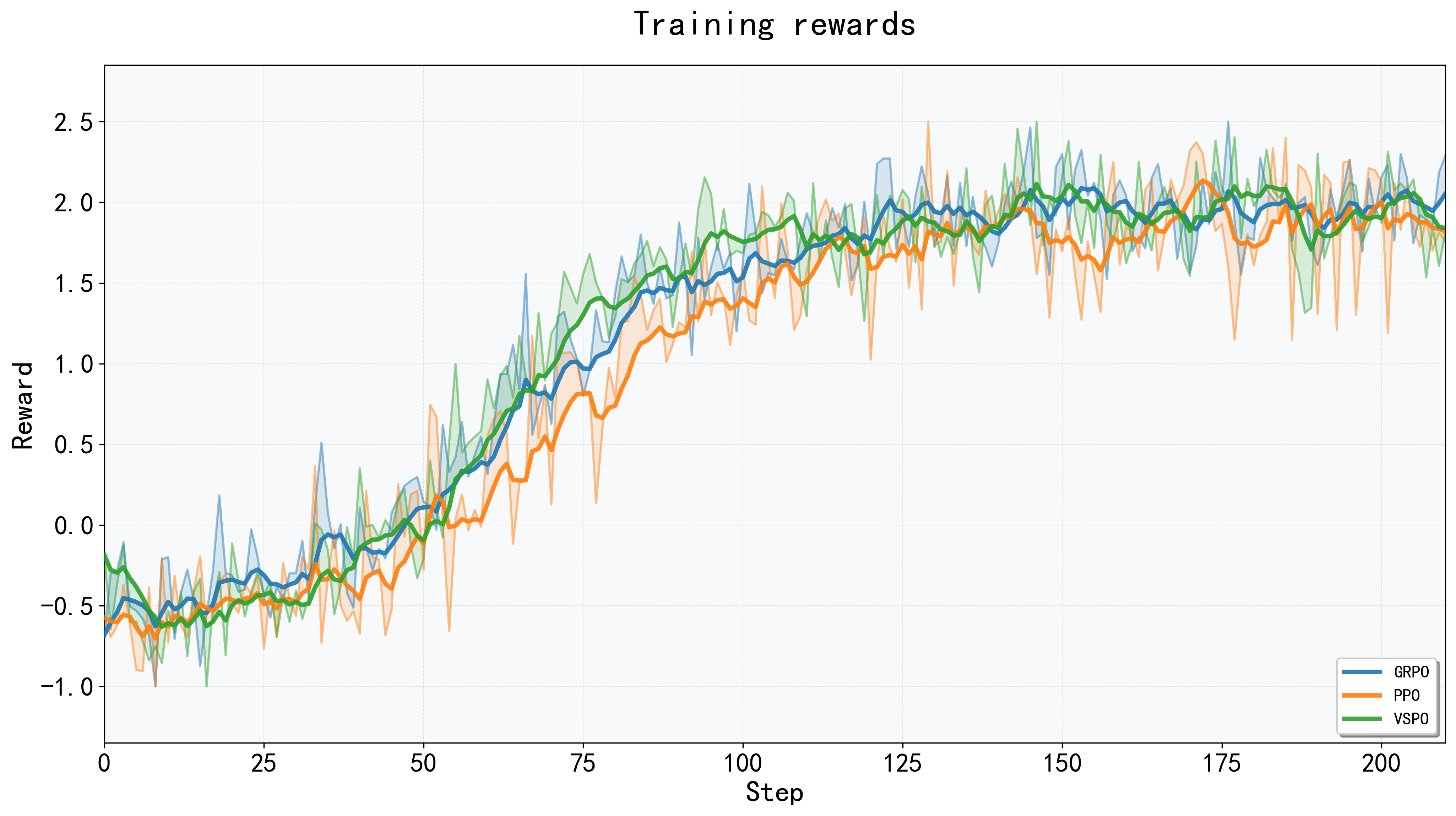}
    \includegraphics[width=0.9\linewidth]{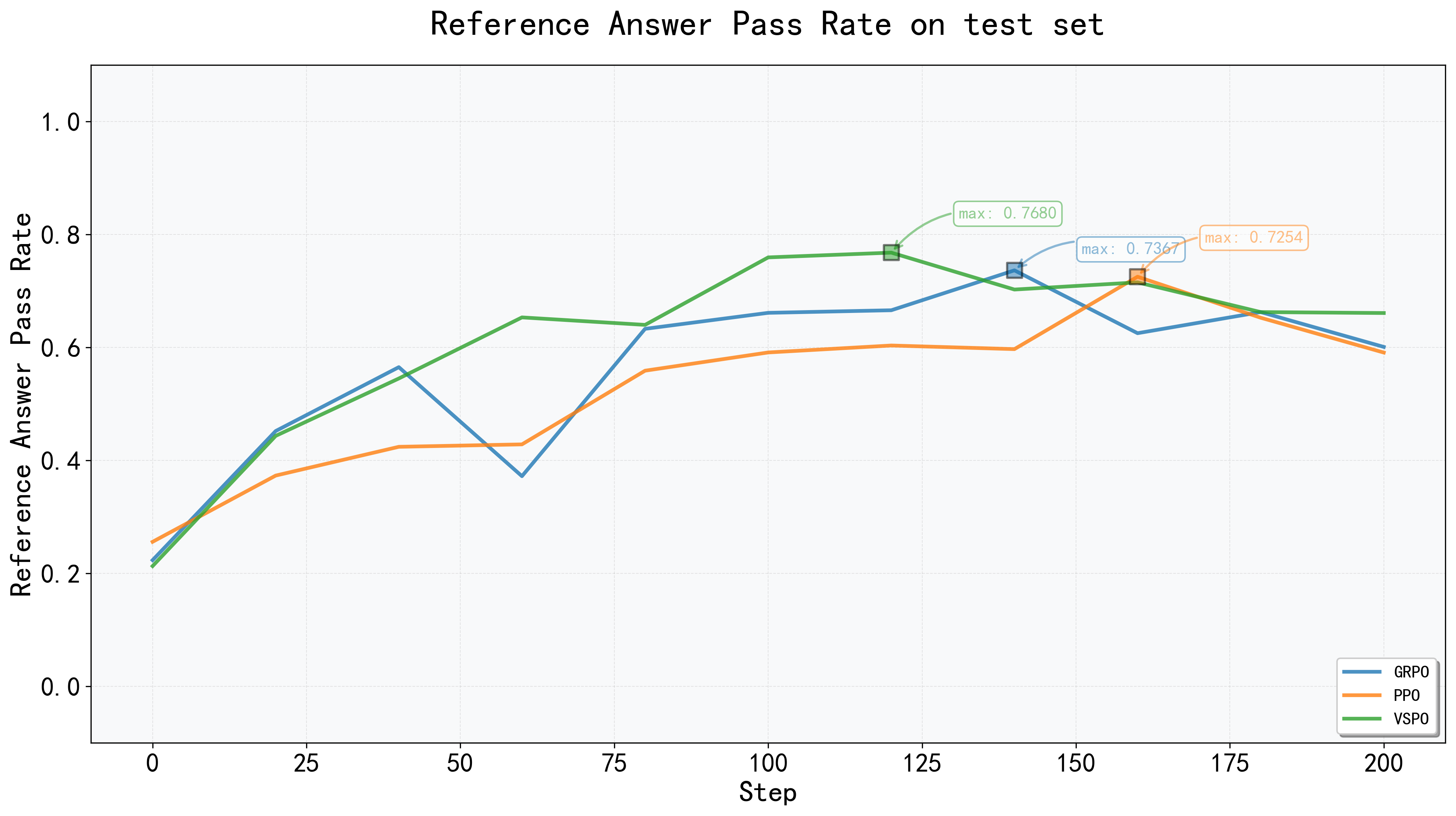}
    \caption{Reward curves on the training and test set metric(Reference Answer Pass Rate) for RL algorithm in long-form QA.}
    \label{fig:po}
\end{figure}

The training reward and test set metric of both RL algorithm are show in Firure \ref{fig:po}. 

VSPO achieved the best performance based on rewards on the train set and reference answer pass rate on the test sets. As the number of training steps increases, PPO gradually converges to the same level as VSPO and GRPO on training set. 

Among all RL algorithms, VSPO was the first to receive the best test metric performance, this indicates that filling in tasks that cannot be helpful for learning through value-based sampling makes the model more focused on high-value tasks, which helps to increase convergence speed and achieve better performance.

\subsection{The Effect of $T$ and $\alpha$ of VSPO}
We analyze the effect of $T$ and $\alpha$ of VSPO. Based on equation \ref{eq:tmp_sample}, We can analyze that $T$ controlling the sharpness of value distribution. If $T$ tends towards 0, the value-based sampling will be concentrated in high learning value tasks. If $T$ tends towards positive infinity, the value-based sampling will become flat and uniform and weaken the gap in learning value of tasks. 

$\alpha$ controlls the maximum amplitude of sample update. The larger the value of $\alpha$, the more radical updates are allowed, while the smaller the value, the more conservative updates are allowed.

\begin{figure}[htb]
    \centering
    % 第一张图
    \includegraphics[width=0.9\linewidth]{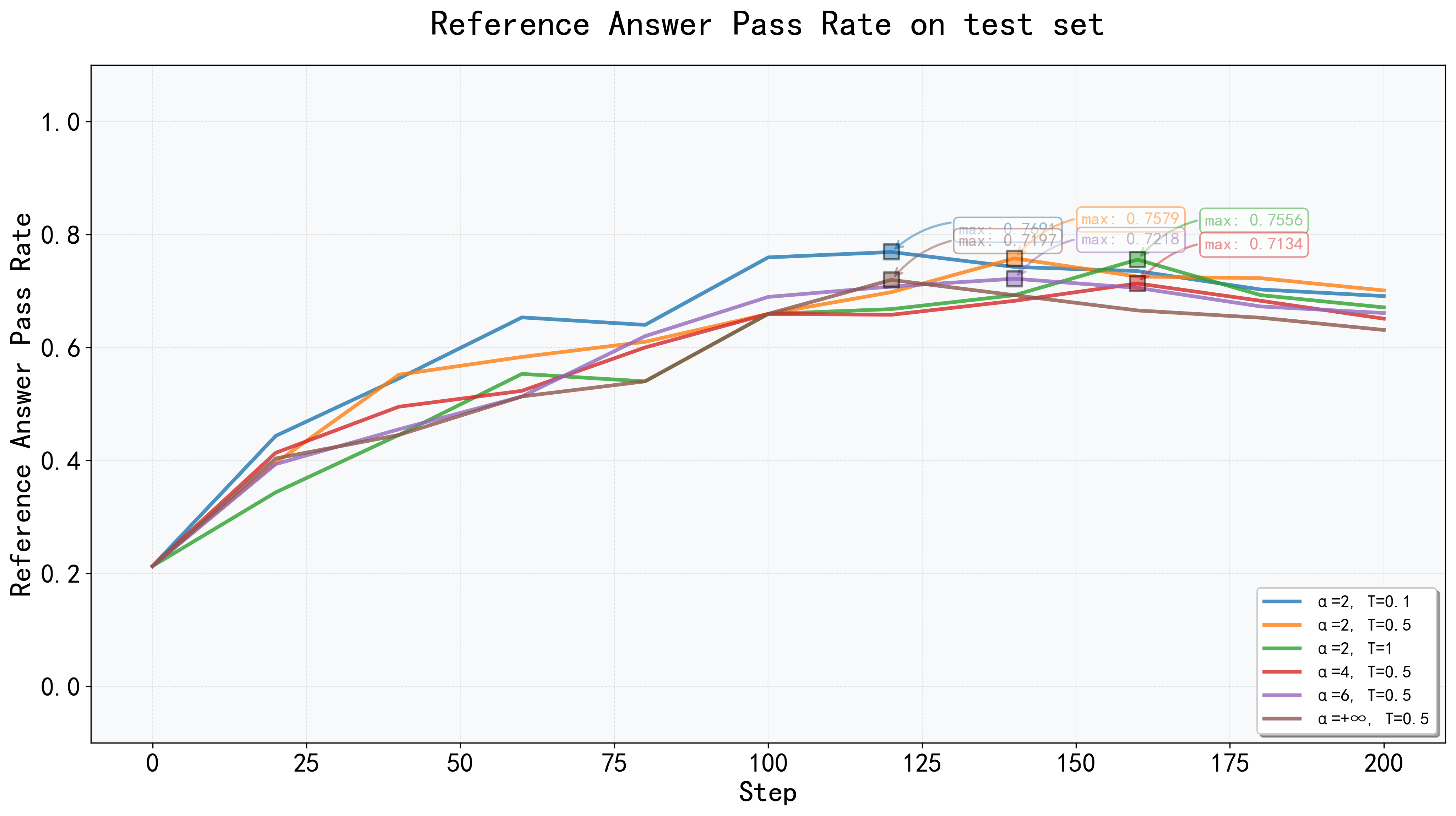}
    \caption{Reference Answer Pass Rate of test set of different $\alpha$ and $T$.}
    \label{fig:at_vspo_long_test}
\end{figure}

The results of different $\alpha$ and $T$ are shown in Table \ref{tab:at_vspo_long} and the reference answer pass rate curve of test set is shown in Figure \ref{fig:at_vspo_long_test}.

\begin{table}[ht!] % 这里改为 table 而不是 table*
\centering
\setlength{\tabcolsep}{4pt} % 调整列间距（可选）
\renewcommand{\arraystretch}{1.2} % 调整行高（可选）
\begin{tabular}{l c c c c}
\hline
\multicolumn{5}{c}{\textbf{Qwen3-14B}} \\
Methods & $Q_{simple}$ & $Q_{multiq}$ & $Q_{multim}$ &Avg. \\ 
\hline
\multicolumn{5}{c}{\textbf{Reference Answer Pass Rate}} \\
Untrained & 0.5513 & 0.6999 & 0.5000 & 0.5897 \\ 
$\alpha = 2$, $T = 1$ & 0.7019 & 0.8833 & 0.8571 & 0.7556 \\
$\alpha = 2$, $T = 0.5$ & 0.7276 & 0.8250 & 0.8571 & 0.7579 \\
$\alpha = 2$, $T = 0.1$ & 0.7147  & 0.9000  & 0.8557  & 0.7690 \\
$\alpha = 4$, $T = 0.5$ & 0.6923  & 0.875  & 0.8571  & 0.7134 \\
$\alpha = 6$, $T = 0.5$ & 0.7051  & 0.8750  & 0.8571  & 0.7218 \\
no clip, $T = 0.5$ & 0.6955  & 0.8917  & 0.996  & 0.7197 \\
\\
\hline
\multicolumn{5}{c}{\textbf{Hallucination Pass Rate}} \\
Untrained         & 0.7821 & 0.8167 & 0.7143 & 0.7893 \\ 
$\alpha = 2$, $T = 1$ & 0.8782 & 0.8667 & 0.8658 & 0.8747 \\
$\alpha = 2$, $T = 0.5$ & 0.8846 & 0.8167 & 0.8571 & 0.8655 \\
$\alpha = 2$, $T = 0.1$ & 0.8662  & 0.9000  & 0.8571  & 0.8750 \\
$\alpha = 4$, $T = 0.5$ & 0.8782  & 0.9167  & 0.7143  & 0.8494 \\
$\alpha = 6$, $T = 0.5$ & 0.8526  & 0.8667  & 0.4286  & 0.8177 \\
no clip, $T = 0.5$ & 0.8526  & 0.8333  & 0.8714  & 0.8075 \\
\hline
\end{tabular}
\caption{Results on the test set of different $\alpha$ and $T$ of VSPO for long-form QA.}
\label{tab:at_vspo_long}
\end{table}

From Table \ref{tab:at_vspo_long} and Figure \ref{fig:at_vspo_long_test}, We can analyze that, in VSPO:
\begin{itemize}
    \item When $\alpha$ is fixed, a lower $T$ will bring better results and as $T$ decreases, the best result will be achieved earlier. This is because a lower $T$ allows value-based sampling to be repeated more times on more valuable tasks, which  allows more valuable tasks to have a relatively larger gradient modulus to update themselves.
    
    \item When $T$ is fixed, raising $\alpha$ did not bring any improvement in effectiveness and $\alpha = +\infty$ (means no value smoothing clipping, corresponding to no clip) lead to a decrease in the effect. This is because the main function of $\alpha$ is to limit the changes in advantages in a smooth manner. If the magnitude of $\alpha$'s changes is too large, it may lead to unstable updates and reduce the effectiveness of training.
\end{itemize}

\subsection{The Effect of PRS}
Not limited to the metric results shown in Table \ref{tab:main_prs_long} and Table \ref{tab:main_prs_short} on the test set, we will further analyze the effect of PRS.

\subsubsection{PRS can reduce the proportion of zero advantage tasks in batch.}
\begin{figure}[htb]
    \centering
    % 第一张图
    \includegraphics[width=0.9\linewidth]{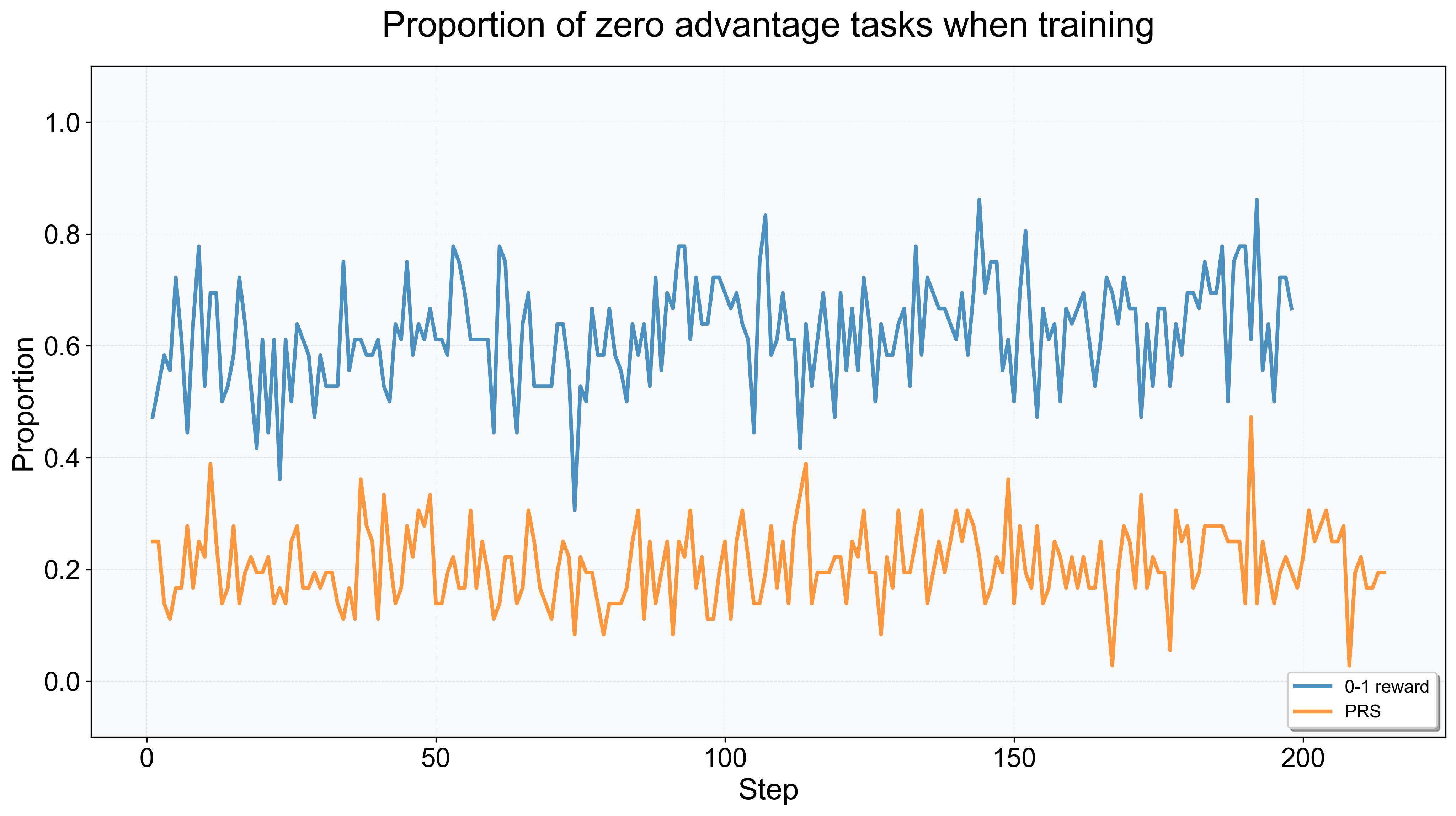}
    \caption{Proportion of zero advantage tasks when training when using 0-1 reward and PRS.}
    \label{fig:zero_adv}
\end{figure}

Figure \ref{fig:zero_adv} shows the proportion of zero advantage tasks at each training step when using 0-1 reward and $PRS_{long}$ (The RL algorithm is GRPO).

We can see that, when using 0-1 reward, about 60\% of tasks for batch in each step do not bring any learning signals. For PRS, based on the summation of various components, it can generate rich partial order, so the proportion of zero advantage tasks is greatly reduced.

\subsubsection{PRS can help policy explore strategies more efficiently.}
\begin{figure}[htb]
    \centering
    % 第一张图
    \includegraphics[width=0.9\linewidth]{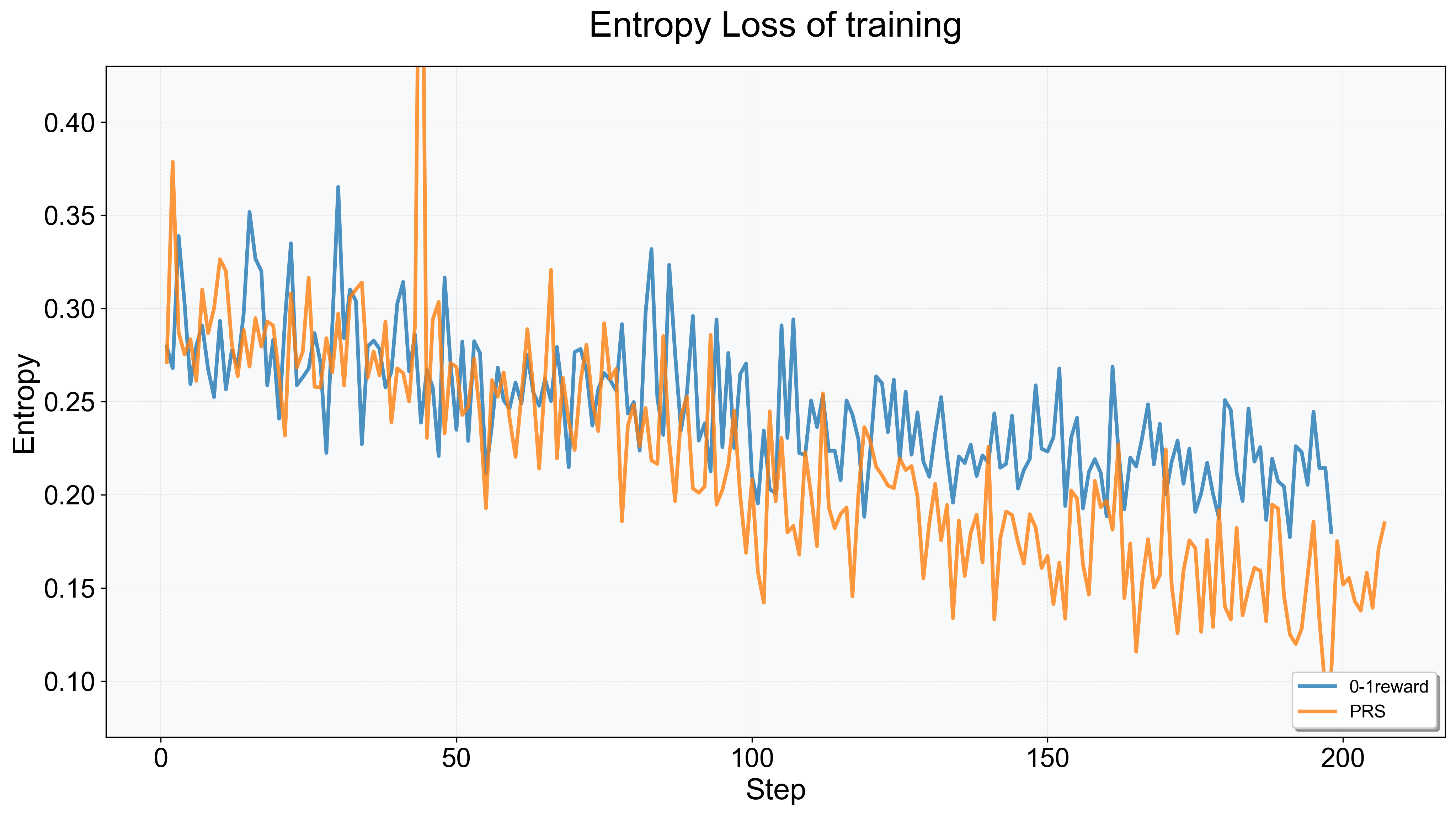}
    \caption{Entropy loss of training using 0-1 reward and PRS.}
    \label{fig:0-1_entro}
\end{figure}

Figure \ref{fig:0-1_entro} shows the entropy loss of VSPO when using 0-1 reward and PRS as training reward. Entropy loss reflects the uncertainty of policy model in selecting the current action. We observe that PRS reduces uncertainty to a lower level earlier than the 0-1 reward and achieves better test set performance. This is not entropy collapse (which would manifest as degraded test metrics), but rather efficient convergence. Therefore, we infer that PRS, as a curriculum-learning-inspired reward design, effectively guides the policy's exploration and enables faster convergence to high-quality strategies. In contrast, the 0-1 reward leads to inefficient exploration under sustained high uncertainty.

\subsubsection{PRS has better improvement on weaker models}
Adding Qwen3-8B to long-form QA and Qwen2.5-7B-Instruct to short-form QA, the results comparing PRS with 0-1 reward are presented in Table \ref{tab:8b_prs_long} for long-form QA and Table \ref{tab:8b_prs_short} for short-form QA.

In long-form OA, $PRS_{long}$ has increased reference answer pass rate by $1.27\%$ ($0.62\%$ for Qwen3-14b) and hallucination Pass Rate by $22.01\%$ ($1.7\%$ for Qwen3-14b) compared to 0-1 reward.
In short-form QA, $PRS_{short}$ has increased exact match score by $2.74\%$ compared to 0-1 reward ($1.08\%$ for Qwen2.5-7b-Instruct).
It can be seen that PRS brings more improvements to weaker models. We believe that the reason is that weak models have weaker exploration ability in the process of RL compared to strong models, so the lack of a guiding reward function (such as 0-1 reward) is a difficult learning signal for weak models. On the contrary, PRS based on curriculum learning gradually decomposes complex tasks through reward settings, which helps weak models gradually explore and improve strategy performance.

\begin{table}[tbp] % 这里改为 table 而不是 table*
\centering
\setlength{\tabcolsep}{4pt} % 调整列间距（可选）
\renewcommand{\arraystretch}{1.2} % 调整行高（可选）
\begin{tabular}{l c c c c}
\hline
\multicolumn{5}{c}{\textbf{Qwen3-8B}} \\
Methods & $Q_{simple}$ & $Q_{multiq}$ & $Q_{multim}$ &Avg. \\ 
\hline
\multicolumn{5}{c}{\textbf{Reference Answer Pass Rate}} \\
Untrained & 0.5353 & 0.1083 & 0.3571 & 0.4148 \\ 
GRPO + 0-1 reward & 0.6731 & 0.1000 & 0.6429 & 0.5180 \\ 
GRPO + $PRS_{long}$        & 0.6506 & 0.1917 & 0.5714 & 0.5246 \\
\hline
\multicolumn{5}{c}{\textbf{Hallucination Pass Rate}} \\
Untrained         & 0.7372 & 0.1000 & 0.7143 & 0.5650 \\ 
GRPO + 0-1 reward & 0.8974 & 0.1167 & 0.8971 & 0.6873 \\ 
GRPO + PRS        & 0.8654 & 0.7667 & 0.8571 & 0.8386\\
\hline
\end{tabular}
\caption{Results on the test set of 0-1 reward and $PRS_{long}$ for long-form QA.}
\label{tab:8b_prs_long}
\end{table}

\begin{table*}[tbp] % 跨双栏，如果不需要可改为 table
\centering
\setlength{\tabcolsep}{4pt} % 调整列间距（可选）
\renewcommand{\arraystretch}{1.2} % 调整行高（可选）
\begin{tabular}{c c c c c c c c c} % 9列，居中对齐
\hline
\multicolumn{9}{c}{\textbf{Qwen2.5-7B-Instruct}} \\
Methods & NQ & TriviaQA & PopQA & HotpotQA & 2wiki & Musique & Bamboogle & Avg. \\ \hline
\multicolumn{9}{c}{\textbf{Exact Match (EM)}} \\
Untrained & 0.2770 & 0.4872 & 0.3289 & 0.2797 & 0.2411 & 0.1037 & 0.2320 & 0.3210 \\ 
GRPO + 0-1 reward & 0.3500 & 0.5438 & 0.3754 & 0.3600 & 0.3095 & 0.1286 & 0.2996 & 0.3855 \\ 
GRPO + $PRS_{short}$ & 0.3573 & 0.5623 & 0.3815 & 0.3757 & 0.3174 & 0.0996 & 0.3333 & 0.3897 \\ \hline
\end{tabular}
\caption{Results on the test set of 0-1 reward and $PRS_{short}$ for short-form QA.}
\label{tab:8b_prs_short}
\end{table*}

\subsection{Ablation Analysis of Value-based Sampling and Value Smoothing Clipping}
To evaluate the effectiveness of value-based sampling and value smoothing clipping of VSPO, we conduct experiments using no sample without clip (equivalent to GRPO), random sample without clip, value-based sample without clip, random sample with clip and value-based sample with clip (equivalent to VSPO), respectively. The result on the test set is shown in Table \ref{tb:vesper_ab}. Table \ref{tb:vesper_ab} demonstrates that removing either value-based sampling or value smoothing clipping significantly degrades performance, confirming the necessity of both components. Notably, any sampling strategy without clipping leads to a decrease in performance, while value-based sampling with clipping achieves the best performance.

\begin{table}[ht!] % 这里改为 table 而不是 table*
\centering
\setlength{\tabcolsep}{4pt} % 调整列间距（可选）
\renewcommand{\arraystretch}{1.2} % 调整行高（可选）
\begin{tabular}{l c c c c}
\hline
\multicolumn{5}{c}{\textbf{Qwen3-14B}} \\
Methods & $Q_{simple}$ & $Q_{multiq}$ & $Q_{multim}$ &Avg. \\ 
\hline
\multicolumn{5}{c}{\textbf{Reference Answer Pass Rate}} \\
\makecell{w/o sample \\ w/o clip (GRPO)} & 0.7144 & 0.8183 & 0.7150 & 0.7300 \\ 
\makecell{random sample \\ w/o clip} & 0.4120 & 0.1600 & 0.5400 & 0.3100 \\
\makecell{value-based sample \\ w/o clip} & 0.6955 & 0.8917 & 0.9960 & 0.7197 \\
\makecell{random sample \\ w clip} & 0.6322  & 0.5790  & 0.7150  & 0.6335 \\
\makecell{value-based sample \\ w clip (VSPO)} & 0.7147  & 0.9000  & 0.8557  & 0.7690 \\
\\
\hline
\multicolumn{5}{c}{\textbf{Hallucination Pass Rate}} \\
\makecell{w/o sample \\ w/o clip (GRPO)} & 0.8982 & 0.9333 & 0.9000 & 0.9077 \\ 
\makecell{random sample \\ w/o clip} & 0.5368 & 0.2300 & 0.5734 & 0.4149 \\
\makecell{value-based sample \\ w/o clip} & 0.8526 & 0.8333 & 0.8714 & 0.8075 \\
\makecell{random sample \\ w clip} & 0.7950  & 0.6333  & 0.8876  & 0.8256 \\
\makecell{value-based sample \\ w clip (VSPO)} & 0.8662  & 0.9000  & 0.8571  & 0.8750 \\
\hline
\end{tabular}
\caption{Results on the test set of different ablation settings for long-form QA.}
\label{tb:vesper_ab}
\end{table}

\begin{figure}[htb]
    \centering
    % 第一张图：KL Loss
    \includegraphics[width=0.9\linewidth]{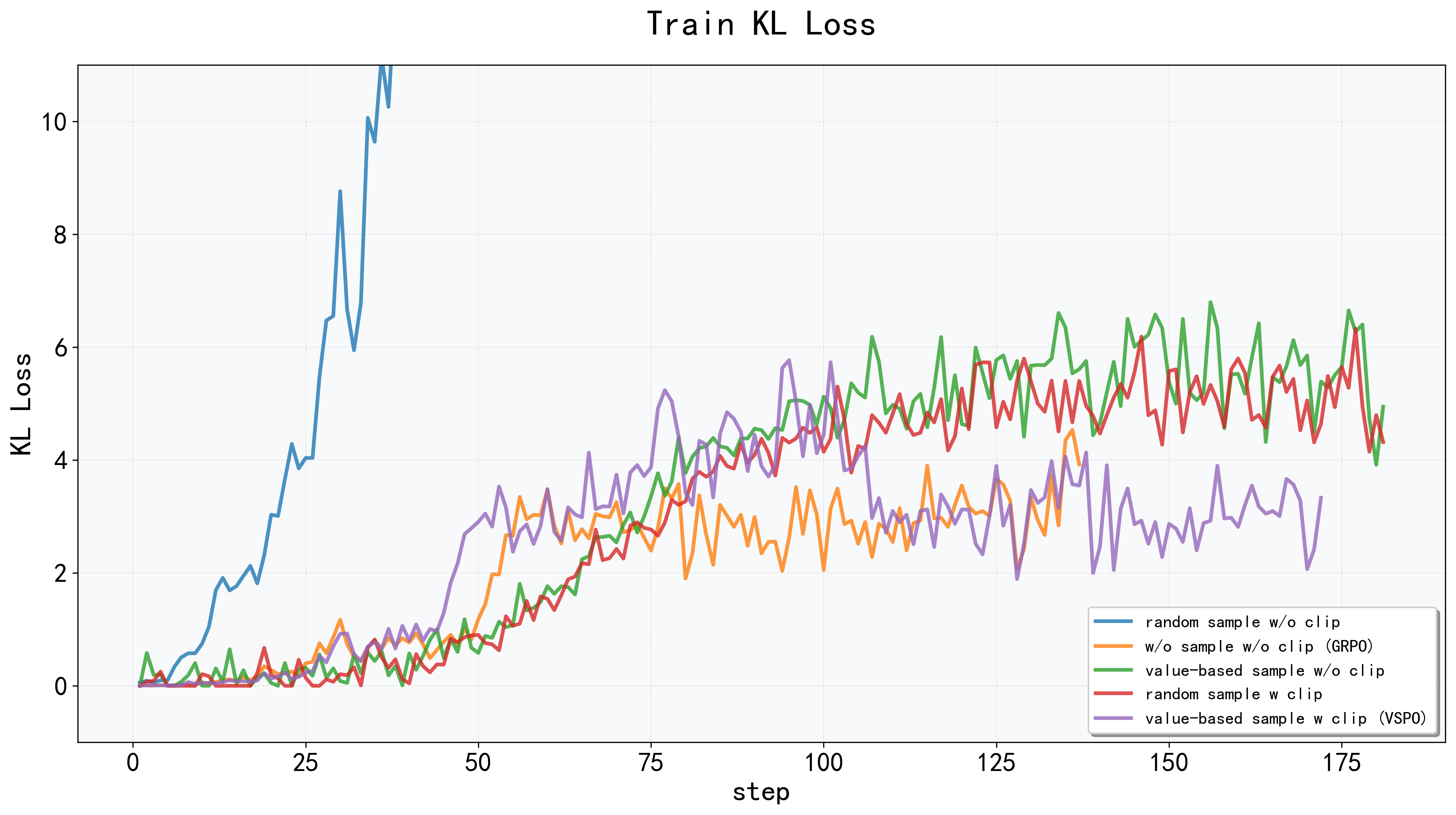}
    \caption*{(a) KL Loss} % 子标题（可选）
    \vspace{0.5em}

    % 第二张图：Reward on eval set
    \includegraphics[width=0.9\linewidth]{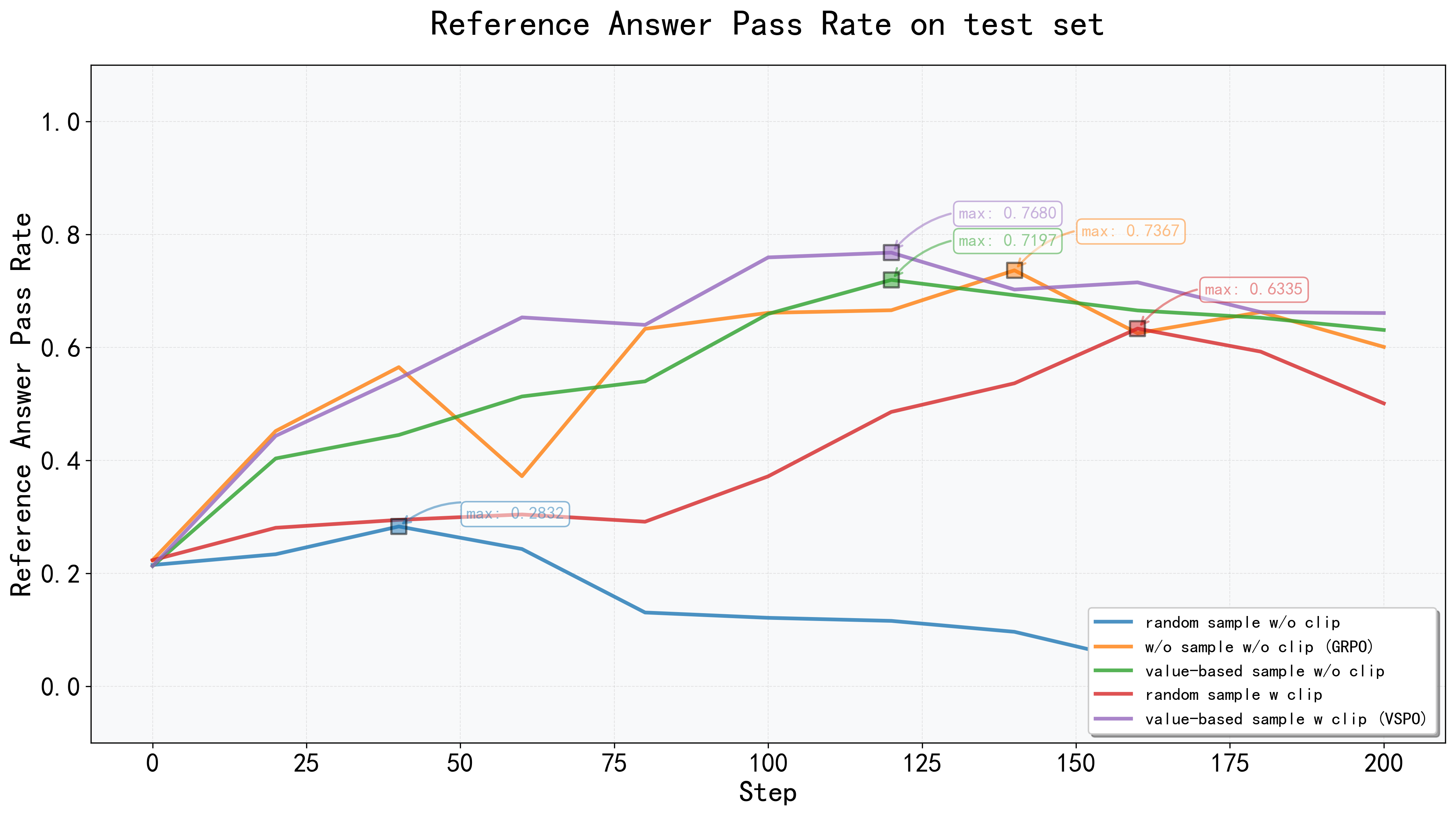}
    \caption*{(b) Reference Answer Pass Rate on test set}
    \vspace{0.5em}

    % % 第三张图：Reward on train set
    % \includegraphics[width=0.9\linewidth]{fig/vesper_ab/vesper_ab_train_reward.png}
    % \caption*{(c) Reward on train set}

    \caption{The KL Loss and Reference Answer Pass Rate on test set.}
    \label{fig:vesper_ab}
\end{figure}

We further analyze the impact of value-based sampling and value smoothing clipping through KL divergence on train set and reference answer pass rate on test set. Figure \ref{fig:vesper_ab} presents these metrics across different configurations. Two key observations emerge: 

(1) \textbf{Value smoothing clipping stabilizes training.} When using random sampling, without clipping causes random sampling abnormally high KL divergence and low metric on test set, which leading to extremely unstable training. With clip can change the low performance of random sampling to some extent. When using value-based sampling, without clipping will have a larger KL loss and lower reference answer pass rate on test set than with clip. The above information can verify the effectiveness of value smoothing clipping. By limiting the changes in advantages in a smooth
manner, value smoothing clipping stabilizes training while improving sampling performance.

(2) \textbf{Value-based sampling is necessary for performance gains.} When using value smoothing clipping, value-based sampling have a higher reference answer pass rate on test set on random sampling. When not using value smoothing clipping, random sampling have abnormally high KL divergence and lower metric on test set than value-based sampling. This confirming the necessity of intelligent sample selection based on learning value.

\section{Conclusions}
In this work, we formalized the Tool-Integrated Reasoning (TIR) problem and proposed two complementary techniques to improve Agentic RL in complex reasoning tasks: \emph{Progressive Reward Shaping} (PRS) and \emph{Value-based Sampling Policy Optimization} (VSPO). 

PRS introduces a curriculum-based reward design that provides dense, stage-wise learning signals, enabling models to first master essential capabilities—such as generating parseable outputs and properly formatted responses—before optimizing for challenging objectives like answer quality and factual grounding. We instantiated PRS in both short-form and long-form QA, with the short-form version incorporating a length-aware BLEU to avoid unfair penalization of correct short answers, and the long-form version incorporating an LLM-as-a-Judge signal to address reward hacking in complex outputs.

To address gradient degradation in GRPO when rollouts have identical rewards, we developed VSPO, which uses value-based sample to prioritize updates on tasks with higher learning value and applies value smoothing clipping to stabilize policy gradients. This combination ensures that policy updates focus on the most informative samples while preventing destabilization from repeated high-value prompts.

Extensive experiments on multiple short-form and long-form QA benchmarks demonstrate that PRS consistently outperforms traditional binary rewards, and VSPO achieves more stable training, better sample efficiency and superior performance across policy optimization algorithms. Overall, this work provides a generalizable reward shaping and policy optimization in Agentic RL with TIR, applicable beyond QA tasks to any multi-step reasoning scenarios.

% \begin{table*}[tbp] % h=here, t=top, b=bottom, p=page float
% \centering          % 表格居中
% \begin{tabular}{c c c c c}
% \hline
%  & $Q_{simple}$ & $Q_{multiq}$ & $Q_{multim}$   \\ \hline
% \makecell{w/o sample \\  w/o clip} & 0.700 & 0.575 & 0.475  \\ \hline
% \makecell{random sample \\ w/o clip} & 0.46 & 0.16 & 0.1   \\ \hline
% \makecell{value-based sample \\ w/o clip} & 0.45 & 0.175 & 0.3  \\ \hline
% \makecell{random sample \\ w/ clip} & 0.5 & 0.3 & 0.025  \\ \hline
% \makecell{value-based sample \\ w/ clip} & 0.7125 & 0.725 & 0.55  \\ 
% \hline
% \end{tabular}
% \caption{Results on the test set of different ablation setting.}
% \label{tb:vesper_ab}
% \end{table*}

\clearpage

\bibliography{aaai2026}

\appendix
\lstset{
    basicstyle=\ttfamily\scriptsize,
    breaklines=true,
    frame=single,
    numbers=none,
    backgroundcolor=\color{gray!5},
    tabsize=2,
    literate={’}{{'}}1
             {“}{{"}}1
             {”}{{"}}1
}

% \section{Implementation details}\label{sec:detail}
% We use Qwen3-235B-A22B-Instruct-2507~\cite{qwen3technicalreport} as an LLM judge for hallucination in $PRS_{long}$.

\section{Prompt Template for QA Agent}
\begin{lstlisting}
Answer the given question. You must conduct reasoning inside <reasoning> and </reasoning> first every time you get new information. After reasoning, if you find you lack some knowledge, you can call wiki_search, and it will return the top searched results. You can search as many times as you want. If you find no further external knowledge needed, you can directly provide the answer inside <answer> and </answer> without detailed illustrations. For example, <answer> xxx </answer>.
Question: {query}
\end{lstlisting}

\section{Prompt Template for LLM-as-aJudge Evaluation}\label{sec:judge_pt}
\begin{lstlisting}
Role and Background
You are an experienced AI capability evaluation expert, with rigorous logic and keen observation skills, highly adept at assessing whether AI responses meet the requirements.
Now, an online travel platform has launched a Q&A Agent that provides answers to merchants' questions.

Your Task
You will receive the following information:
[Merchant's Question]: The question raised by the merchant.
[Q&A Agent's Response]: The answer provided by the Q&A Agent.
[Reference Answer]: The standard answer corresponding to the merchant's question.
You need to, strictly based on the [Evaluation Criteria], give an evaluation of the [Q&A Agent's Response].
You must strictly follow the internal step-by-step thinking process, comprehensively analyze and evaluate the [Q&A Agent's Response] based on the provided relevant information, and output the evaluation result in the specified format.

Your Internal Thinking Process (Please carry out internally, do not output)
Follow the steps below in your mind, analyzing step by step:

Analyze [Evaluation Criteria]
Carefully read the Evaluation Criteria, understand the core evaluation question and the requirements of each criterion.

Analyze [Merchant's Question]
Analyze and understand the Merchant's Question. Some questions may contain multiple sub-questions, others may include images (in URL form). If there are multiple questions, you need to understand the meaning of each.

Analyze [Q&A Agent's Response]
Analyze and understand the Q&A Agent's Response. If the Merchant's Question contains multiple sub-questions, the Q&A Agent's Response may contain multiple answers. You need to understand each answer, extract the information inside, determine which question it answers, and check if the model responded to all the questions.

Analyze [Reference Answer]
Analyze and understand the Reference Answer. If the Merchant’s Question contains multiple sub-questions, the Reference Answer will have answers to each question, and sometimes the order may differ from the order of questions in the Merchant’s Question. You need to first understand each answer, extract the information inside, match it with the corresponding question, and know the reference answer for each.

Perform Evaluation
Based on the above analysis results, use the [Evaluation Criteria] and [Reference Answer] to analyze and evaluate the [Q&A Agent’s Response].

Evaluation Criteria
Core evaluation question: Does the [Q&A Agent’s Response] match the [Reference Answer]?
Evaluation criteria:

Match: The Agent’s response must be semantically consistent with the Reference Answer. If the Merchant’s Question contains multiple sub-questions, the Q&A Agent’s Response must include an answer to each question, and each must be semantically consistent with the corresponding Reference Answer.
Mismatch: If the above “Match” criterion is not met, then it is judged as “Mismatch”. That is, the Q&A Agent’s Response has significant deviations from the Reference Answer. For example, when the merchant asks multiple questions, the Q&A Agent’s Response only answers some of them, etc.
Output Format Requirement
You must strictly follow the format below when outputting, and directly provide the evaluation rating without any extra explanation or comments.
<Evaluation Rating>:
\end{lstlisting}

% Check whether the conference requires a reproducibility checklist to be included in the paper.
% If so, you can uncomment the following line and ajust the path to include it.
% \input{../../ReproducibilityChecklist/LaTeX/ReproducibilityChecklist.tex}

\end{document}